\documentclass[sigconf]{acmart}
\AtBeginDocument{%
  }

\copyrightyear{2025}
\acmYear{2025}
\setcopyright{cc}
\setcctype{by}
\acmConference[CIKM '25]{Proceedings of the 34th ACM International Conference on Information and Knowledge Management}{November 10--14, 2025}{Seoul, Republic of Korea}
\acmBooktitle{Proceedings of the 34th ACM International Conference on Information and Knowledge Management (CIKM '25), November 10--14, 2025, Seoul, Republic of Korea}\acmDOI{10.1145/3746252.3761103}
\acmISBN{979-8-4007-2040-6/2025/11}

\usepackage{graphicx}
\usepackage{multirow}
\usepackage{subcaption}
\usepackage{booktabs}
\usepackage{array}
\usepackage{enumitem}
\usepackage{siunitx}
\usepackage{makecell}
\usepackage{amsmath}
\usepackage{pifont}
\usepackage{algorithm}
\usepackage{algpseudocode}
\newtheorem{definition}{Definition}
\newtheorem{theorem}{Theorem}

\makeatletter
\newcommand*{\indep}{%
	\mathbin{%
		\mathpalette{\@indep}{}%
	}%
}
\newcommand*{\nindep}{%
	\mathbin{
		\mathpalette{\@indep}{\not}
	}%
}
\newcommand*{\@indep}[2]{%
	\sbox0{$#1\perp\m@th$}
	\sbox2{$#1=$}
	\sbox4{$#1\vcenter{}$}
	\rlap{\copy0}
	\dimen@=\dimexpr\ht2-\ht4-.2pt\relax
	\kern\dimen@
	{#2}%
	\kern\dimen@
	\copy0 %
}

\begin{document}

\title[CFD-Prompting: Unbiased Reasoning in LLMs]{Unbiased Reasoning for Knowledge-Intensive Tasks in Large Language Models via Conditional Front-Door Adjustment}

\author{Bo Zhao}
\authornote{Both authors contributed equally to this research.}
\email{zbo@webmail.hzau.edu.cn}
\orcid{0009-0007-3114-4452}
\affiliation{%
	\institution{Huazhong Agricultural University}
	\city{Wuhan}
	\country{China}
}

\author{Yinghao Zhang}
\authornotemark[1]
\email{yhzhang@mail.hzau.edu.cn}
\orcid{0000-0001-6084-3278}
\affiliation{%
	\institution{Huazhong Agricultural University}
	\city{Wuhan}
	\country{China}
}

\author{Ziqi Xu}
\authornote{Corresponding author.}
\email{ziqi.xu@rmit.edu.au}
\orcid{0000-0003-1748-5801}
\affiliation{%
	\institution{RMIT University}
	\city{Melbourne}
	\country{Australia}
}

\author{Yongli Ren}
\email{yongli.ren@rmit.edu.au}
\orcid{0000-0002-3137-9653}
\affiliation{%
	\institution{RMIT University}
	\city{Melbourne}
	\country{Australia}
}

\author{Xiuzhen Zhang}
\email{xiuzhen.zhang@rmit.edu.au}
\orcid{0000-0001-5558-3790}
\affiliation{%
	\institution{RMIT University}
	\city{Melbourne}
	\country{Australia}
}

\author{Renqiang Luo}
\email{lrenqiang@jlu.edu.cn}
\orcid{0009-0002-8313-9835}
\affiliation{%
	\institution{Jilin University}
	\city{Changchun}
	\country{China}
}

\author{Zaiwen Feng}
\authornotemark[2]
\email{zaiwen.feng@mail.hzau.edu.cn}
\orcid{0000-0003-1618-3553}
\affiliation{%
	\institution{Huazhong Agricultural University}
	\city{Wuhan}
	\country{China}
}

\author{Feng Xia}
\email{f.xia@ieee.org}
\orcid{0000-0002-8324-1859}
\affiliation{%
	\institution{RMIT University}
	\city{Melbourne}
	\country{Australia}
}

\renewcommand{\shortauthors}{Bo Zhao et al.}

\begin{abstract}
	Large Language Models (LLMs) have shown impressive capabilities in natural language processing but still struggle to perform well on knowledge-intensive tasks that require deep reasoning and the integration of external knowledge. Although methods such as Retrieval-Augmented Generation (RAG) and Chain-of-Thought (CoT) have been proposed to enhance LLMs with external knowledge, they still suffer from internal bias in LLMs, which often leads to incorrect answers. In this paper, we propose a novel causal prompting framework, \underline{C}onditional \underline{F}ront-\underline{D}oor Prompting (CFD-Prompting), which enables the unbiased estimation of the causal effect between the query and the answer, conditional on external knowledge, while mitigating internal bias. By constructing counterfactual external knowledge, our framework simulates how the query behaves under varying contexts, addressing the challenge that the query is fixed and is not amenable to direct causal intervention. Compared to the standard front-door adjustment, the conditional variant operates under weaker assumptions, enhancing both robustness and generalisability of the reasoning process. Extensive experiments across multiple LLMs and benchmark datasets demonstrate that CFD-Prompting significantly outperforms existing baselines in both accuracy and robustness. The source code and case study are available at: \url{https://github.com/zbb79/CFD-Prompting}.
\end{abstract}


%

\maketitle

\section{Introduction}
\begin{figure}[t]
	\centering
	\includegraphics[width=0.44\textwidth]{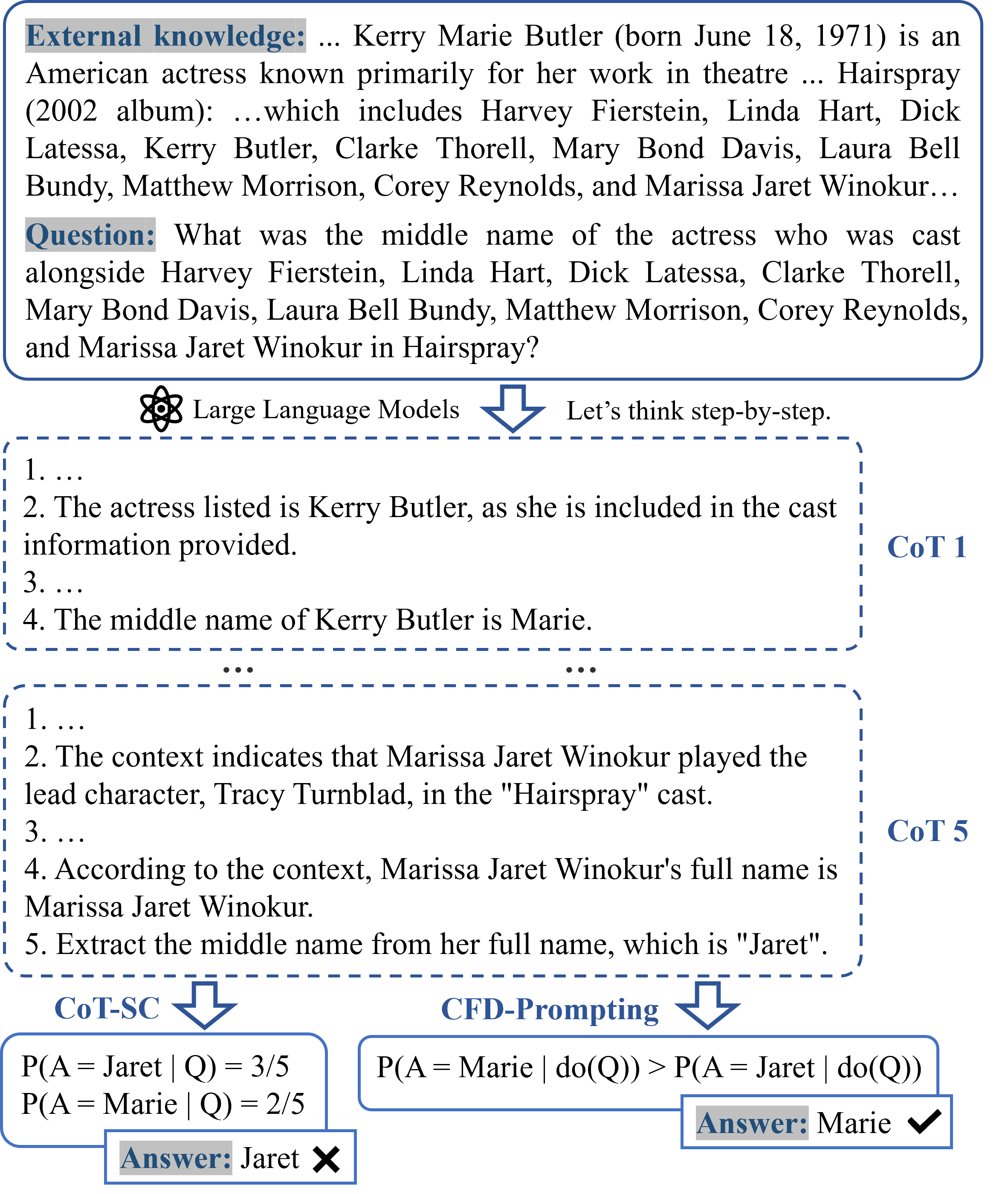}
	\caption{An example illustrating internal bias in LLMs. CoT-SC~\cite{DBLP:conf/iclr/0002WSLCNCZ23} selects the most frequent answer among sampled CoTs, which results in incorrect answer. In contrast, CFD-Prompting selects the answer with the highest causal effect and yields the correct result. This example is taken from a response by GPT-3.5 Turbo on HotpotQA.}
	\label{fig:figure1}
\end{figure}

In recent years, Large Language Models (LLMs) have achieved remarkable progress in natural language processing. By pre-training on massive text corpora, they have demonstrated impressive capabilities in language understanding and generation. Techniques such as in-context learning~\cite{DBLP:conf/nips/BrownMRSKDNSSAA20} and Chain-of-Thought (CoT)~\cite{DBLP:conf/nips/Wei0SBIXCLZ22} have enabled LLMs to make significant advancements in tasks such as question answering and behaviour simulation~\cite{Ma0RHC25}. However, LLMs still face critical challenges in knowledge-intensive tasks, as accurate answers often require specific information that falls outside the distribution of their internal knowledge~\cite{YuanCCGZCJL023}.

To efficiently access external knowledge, relying solely on fine-tuning incurs significant computational costs and is further constrained by limited timeliness~\cite{DBLP:journals/corr/abs-2309-10313}. A promising alternative is incorporating external knowledge directly into prompts~\cite{ZhaoLJQB23}, for example, through Retrieval-Augmented Generation (RAG)~\cite{DBLP:conf/nips/LewisPPPKGKLYR020} or knowledge graphs~\cite{DBLP:journals/corr/abs-2410-14211}, which enable LLMs to access more comprehensive and up-to-date information. However, simply injecting external knowledge into prompts does not guarantee that LLMs can identify and utilise relevant information~\cite{ShiCMSDCSZ23}. Recent studies further find that internal bias in LLMs can lead to spurious correlations with the query, thereby preventing the models from effectively leveraging external information to generate accurate answers~\cite{MallenAZDKH23}.

To address this challenge, CoT self-consistency (CoT-SC)~\cite{DBLP:conf/iclr/0002WSLCNCZ23} improves reasoning by sampling multiple CoTs and selecting the most frequent answer. While effective against random errors, it fails to correct spurious correlations caused by internal bias in LLMs. As a stronger alternative, causality-based prompting methods estimate the causal effect of either the query or the CoT on the answer to select more reliable reasoning paths~\cite{DBLP:journals/corr/abs-2403-02738,DBLP:conf/acl/Wu0CWRKRM24}. As shown in Figure~\ref{fig:figure1}, we present an illustrative example comparing majority voting with our framework. The latter produces more accurate answers by ranking the causal effect of the query on the answer.

\begin{figure}[t]
	\centering
	\begin{subfigure}[t]{0.12\textwidth}
		\includegraphics[width=\textwidth]{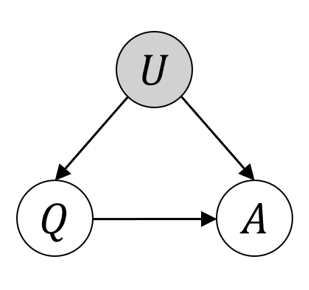}
		\caption{}
		\label{fig:a}
	\end{subfigure}
	\hfill
	\begin{subfigure}[t]{0.17\textwidth}
		\includegraphics[width=\textwidth]{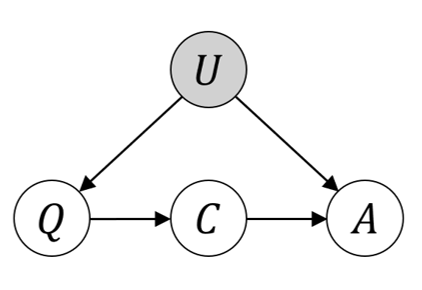}
		\caption{}
		\label{fig:b}
	\end{subfigure}
	\hfill
	\begin{subfigure}[t]{0.17\textwidth}
		\includegraphics[width=\textwidth, clip, trim=0 0 6pt 0]{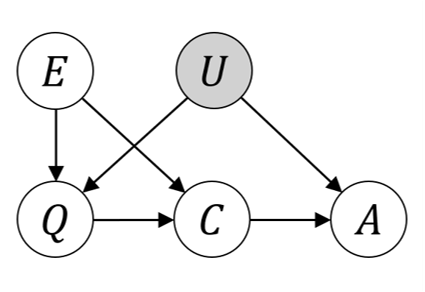}
		\caption{}
		\label{fig:c}
	\end{subfigure}
	\caption{Three SCMs representing reasoning in LLMs: (a) general reasoning without CoTs; (b) CoT and CoT-SC incorporate explicit reasoning, and CP applies the standard front-door adjustment; (c) DeCoT and the proposed CFD-Prompting, which are specifically for knowledge-intensive tasks. Here, $Q$ is the query, $A$ is the answer, $C$ is the CoT, $U$ is the latent confounder, and $E$ is the observed external knowledge.}
	\label{fig:total}
\end{figure}

While prior work has explored both general reasoning and causal-ity-based methods, their differences have not been formally characterised. We introduce a set of Structural Causal Models (SCMs) to clarify the assumptions and limitations of each method. As shown in Figure~\ref{fig:a}, general methods perform direct reasoning without CoTs, where the latent confounder $U$ induces spurious correlations between the query and the answer, often leading to incorrect answers. Figure~\ref{fig:b} illustrates CoT and CoT-SC, which enhance performance through explicit reasoning steps but still suffer from bias due to $U$. Causal prompting (CP) mitigates this bias via the standard front-door adjustment~\cite{DBLP:journals/corr/abs-2403-02738}, but it relies on strong assumptions, namely the absence of any observed confounders that interact with the CoT. To relax these constraints, \citet{DBLP:conf/acl/Wu0CWRKRM24} propose DeCoT for knowledge-intensive tasks (as shown in Figure~\ref{fig:c}), which leverages external knowledge as an instrumental variable to estimate the average causal effect of the CoT on the answer. However, this method provides only a coarse estimate and may overlook fine-grained causal effects between the query and the answer. Thus, there is a clear need for a causal prompting framework that can provide an unbiased estimate of the causal effect of the query on the answer to improve performance on knowledge-intensive tasks.

In this work, we propose the \underline{C}onditional \underline{F}ront-\underline{D}oor Prompting (CFD-Prompting) framework, which leverages conditional front-door adjustment to mitigate internal bias and generate more reliable answers for knowledge-intensive tasks. As a relaxed variant of the standard front-door criterion, it allows interactions between CoTs and external knowledge, making it better suited to such tasks. To implement this, we generate counterfactual external knowledge to simulate causal interventions on the query. CFD-Prompting adopts an encoder-based architecture that does not require access to model logits, enabling compatibility with closed-source LLMs. The contributions of this paper are summarised as follows:

\begin{itemize}[leftmargin=0.6cm]
	\item We present a causal analysis of LLM reasoning using structural causal models, offering a theoretical foundation for de-biasing answers in knowledge-intensive tasks.
	
	\item We propose CFD-Prompting, a general and logit-free causal prompting framework that supports both open-source and closed-source LLMs, and relaxes the assumptions of standard front-door methods by allowing interactions between CoTs and external knowledge.
	
	\item We conduct extensive experiments across multiple LLMs and benchmark datasets, demonstrating that CFD-Prompting consistently outperforms state-of-the-art prompting baselines in both accuracy and robustness.
\end{itemize}

\section{Preliminaries}
We use capital letters to denote variables and lowercase letters to denote their values. Due to space limitations, we refer readers to \cite{Pearl_2009} for the fundamental definitions of causality, including directed acyclic graphs (DAGs), the Markov condition, faithfulness, $d$-separation, and $d$-connection.

\subsection{Structural Causal Model}
The structural causal model (SCM)~\cite{pearl2016causal} formalises causal relationships between variables using a directed acyclic graph (DAG) and a set of structural equations. In a DAG $\mathcal{G} = (\mathcal{V}, \mathcal{E})$, $\mathcal{V}$ denotes the set of nodes (variables), and $\mathcal{E}$ denotes the set of directed edges, where an edge $\mathcal{V}_i \to \mathcal{V}_j$ indicates that $\mathcal{V}_i$ is a direct cause of $\mathcal{V}_j$.

A \emph{path} $\pi$ between nodes $\mathcal{V}_1$ and $\mathcal{V}_n$ is a sequence of distinct nodes $\langle \mathcal{V}_1, \mathcal{V}_2, \dots, \mathcal{V}_n \rangle$ such that each consecutive pair $(\mathcal{V}_i, \mathcal{V}_{i+1})$ is adjacent in the graph. A node $\mathcal{V}$ is said to lie on the path $\pi$ if it appears in the sequence. A path $\pi$ is called \emph{causal} if all edges along it follow the same direction, i.e., $\mathcal{V}_1 \to \mathcal{V}_2 \to \dots \to \mathcal{V}_n$; otherwise, it is referred to as a {non-causal} path.

As illustrated in Figure~\ref{fig:a}, we denote the query as $Q$, which includes both demonstrations and test examples provided to the LLM. The predicted answer generated by the LLM in response to the query is denoted as $A$. Since LLMs generate answers directly based on the given $Q$, we represent the direct causal effect from query to answer as $Q \to A$. However, during the pre-training phases, LLMs may internalise spurious correlations between surface-level patterns and output distributions. These correlations, often inherited from large-scale web corpora or task-specific fine-tuning datasets, can manifest as implicit biases in downstream reasoning~\cite{DBLP:journals/natmi/DingQYWYSHCCCYZWLZCLTLS23, DBLP:conf/acl/AghajanyanGZ20}. 

To model this phenomenon, we introduce an unobservable variable \( U \), which captures the internal bias in LLMs. In this case, although \( Q \to A \) holds as a structural dependency, the true causal relationship is confounded by \( U \), which influences the generation of a reasonable  answer. This is formally captured by the back-door path \( Q \leftarrow U \rightarrow A \), indicating that the observed statistical association between \( Q \) and \( A \) is not purely causal. To estimate the true causal effect of \( Q \) on \( A \), it is necessary to adjust for this confounder \( U \). If the confounder \( U \) is observable, the causal effect could be adjusted using the back-door adjustment formula~\cite{Pearl_2009} as follows:
\begin{equation}
	\centering
	\label{eq:backdoor}
	P(A \mid \mathrm{do}(Q)) = \sum_{u} P(A \mid Q, u) P(u).
\end{equation}

However, in practice, since \( U \) is latent (i.e., unmeasured), alternative strategies such as front-door adjustment are required to obtain an unbiased estimate of the causal effect.

\subsection{Front-door Adjustment}
One prominent approach to addressing unobserved confounders is the standard front-door adjustment~\cite{Pearl_2009}. Unlike the back-door criterion, which requires blocking all back-door paths, the standard front-door criterion isolates the causal pathway through a suitable mediator, even in the presence of unobserved confounders. In the following, we outline the standard front-door criterion and describe how it can be adapted to mitigate internal bias in LLMs.

\begin{definition}[Standard Front-Door Criterion~\citep{Pearl_2009}]
	\label{def:FD}
	A set of variables $Z_{\text{\rm SFD}}$ is said to satisfy the (standard) front-door criterion relative to an ordered pair of variables $(Q,A)$ in a DAG $\mathcal{G}$ if the following conditions hold: (1) $Z_{\text{\rm SFD}}$ intercepts all directed paths from $Q$ to $A$; (2) there is no unblocked back-door path from $Q$ to $Z_{\text{\rm SFD}}$; (3) all back-door paths from $Z_{\text{\rm SFD}}$ to $A$ are blocked by $Q$.
\end{definition}

\begin{theorem}[Standard Front-Door Adjustment~\citep{Pearl_2009}]
	If $Z_{\text{\rm SFD}}$ satisfies the standard front-door criterion relative to $(Q,A)$, then the causal effect of $Q$ on $A$ is identifiable and is given by the following standard front-door adjustment formula:
	\begin{equation}
		\label{eqa:FD2009}
		\begin{aligned}
			P(A|do(Q)) = \sum_{z_{\text{\rm SFD}},~q}^{}P(A\mid q,z_{\text{\rm SFD}})P(z_{\text{\rm SFD}}\mid q)P(q).
		\end{aligned}
	\end{equation}
\end{theorem}

The derivation of Equation~\ref{eqa:FD2009} relies on the rules of do-calculus~\citep{Pearl_2009}, which allows the systematic transformation of expressions involving the $do(\cdot)$ operator into observational probabilities under certain graphical conditions. The rules of do-calculus are as follows:

\begin{theorem}[Rules of $do$-Calculus~\citep{Pearl_2009}]
	\label{do}
	Let $\mathcal{G}$ be the DAG associated with a structural causal model, and let $P(\cdot)$ denote the probability distribution induced by that model. For any disjoint subsets of variables $Q, A, Z$, and $W$, the following rules hold:
	\begin{itemize}[leftmargin=0.4cm]
		\item Rule 1 (Insertion/deletion of observations): $P(A \mid do(A), Z, W) = P(A \mid do(Q), W)$, if $(A \indep Z \mid A, W)$ in $\mathcal{G}_{\overline{Q}}$;
		\item Rule 2 (Action/observation exchange): $P(A \mid do(Q), do(Z), W) = P(A \mid do(Q), Z, W)$, if $(Y \indep Z \mid Q, W)$ in $\mathcal{G}_{\overline{Q}\underline{Z}}$;
		\item Rule 3 (Insertion/deletion of actions): $P(A \mid do(Q), do(Z), W) = P(A \mid do(Q), W)$, if $(A \indep Z \mid Q, W)$ in  $\mathcal{G}_{\overline{Q}, \overline{Z(W)}}$, 
	\end{itemize}
	where $Z(W)$ is the set of nodes in $Z$ that are not ancestors of any node in $W$ in $\mathcal{G}_{\overline{Q}}$. 
\end{theorem}

Here, $\mathcal{G}_{\overline{Q}}$ denotes the DAGs obtained by removing all incoming edges into $Q$, while $\mathcal{G}_{\underline{Q}}$ denotes the graph obtained by removing all outgoing edges from $Q$. This notation generalises to any variable or set of variables, not limited to $Q$.

The standard front-door criterion provides a theoretical foundation for identifying causal effects even in the presence of unobserved confounders. In the context of LLMs, this insight motivates treating the CoT as a valid front-door variable for estimating the causal effect of the query on the answer . As illustrated in Figure~\ref{fig:b}, \( C \) satisfies the conditions of the standard front-door criterion with respect to the causal effect of \( Q \) on \( A \). This allows the causal effect \( P(A \mid \text{do}(Q)) \) to be decomposed into two components: the effect of \( Q \) on \( C \), and the effect of \( C \) on \( A \) conditional on \( Q \). Specifically, the front-door adjustment formula is given by:
\begin{equation}
	\centering
	P(A \mid do(Q)) = \sum_{c} P(c \mid Q) \sum_{q} P(A \mid c, q) P(q).
\end{equation}

CP is a causality-based prompting framework grounded in the SCM shown in Figure~\ref{fig:b}. It assumes no observed variables interact with the CoT, simplifying the DAG to satisfy the standard front-door criterion~\cite{DBLP:journals/corr/abs-2403-02738}. However, this assumption limits its applicability to knowledge-intensive tasks, where external knowledge \( E \) often acts as an observed confounder influencing both the query and the CoT. In such cases, the CoT no longer meets the conditions for a valid front-door variable.

To address the limitations of standard front-door adjustment in knowledge-intensive tasks, we adopt the conditional front-door adjustment, which accounts for observed confounders through conditioning. This motivates our proposed CFD-Prompting framework, which treats the CoT as a conditional front-door variable and incorporates external knowledge to enable unbiased reasoning.

\begin{figure}[t]
	\centering
	\includegraphics[width=0.45\textwidth]{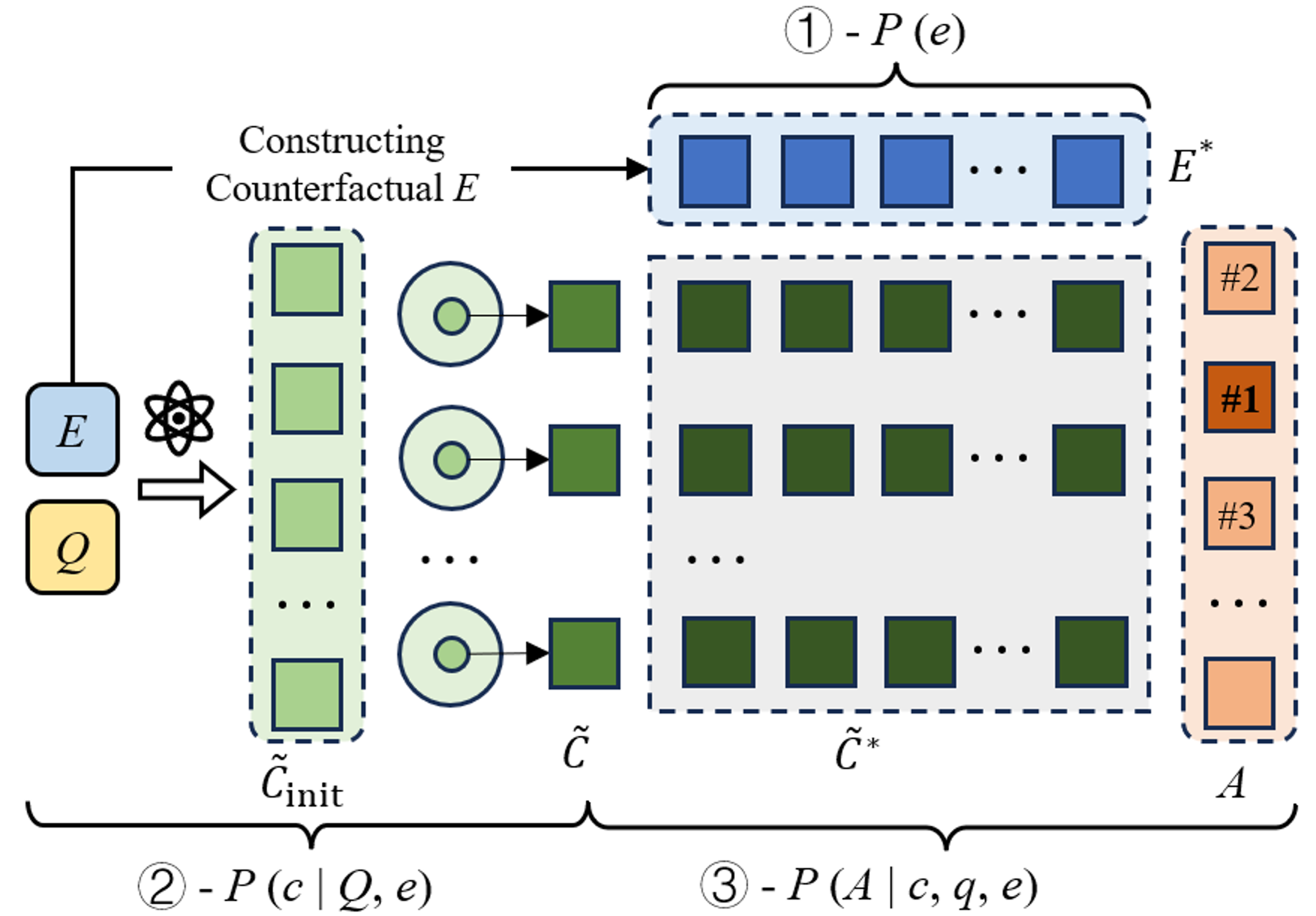} 
	\caption{The overall architecture of CFD-Prompting. $\tilde{C}_{\text{init}}$ denotes the encoded CoT generated by the LLM given the query $Q$ and external knowledge $E$; $\tilde{C}$ is the encoded CoT after applying k-means; $\tilde{C}^*$ represents the encoded CoT generated using the counterfactual variant of external knowledge $E^*$; $\#1$ indicates the final answer selected based on the highest estimated causal effect, i.e., $P(A \mid \text{do}(Q))$.}
	\label{fig:flowchart}
\end{figure}

\section{Method}
In this section, we first outline the task and introduce the notations used throughout the paper. We then present our proposed framework, CFD-Prompting, which decomposes the overall effect into three components and enables unbiased answer selection via causal interventions. The overall architecture of CFD-Prompting is illustrated in Figure~\ref{fig:flowchart}.

\subsection{Task Description}
We consider knowledge-intensive tasks in which an LLM is prompted with a query \( Q \), generates a CoT \( C \), and subsequently produces a final answer \( A \), as illustrated in Figure~\ref{fig:c}. This SCM, inspired by~\cite{DBLP:conf/acl/Wu0CWRKRM24}, assumes that both \( Q \) and \( C \) may be influenced by external knowledge \( E \), while \( U \) represents a latent confounder that biases the estimation of the causal effect of \( Q \) on \( A \).

To address this issue, we adopt a conditional front-door adjustment strategy to obtain an unbiased estimate of \( P(A \mid do(Q)) \). By recovering the unbiased causal effect, we are able to select the answer with the highest causal effect from the query, which we regard as the most reliable or correct answer.

\subsection{Conditional Front-Door Adjustment}
To mitigate the bias introduced by \( U \), we leverage conditional front-door adjustment. The formal criterion is defined as follows: 
\begin{definition}[Conditional Front-Door Criterion~\cite{DBLP:conf/iclr/XuCLL0Y24}]
	\label{def:FD1}
	A set of variables \( Z_{\text{\rm CFD}} \) is said to satisfy the \textit{conditional front-door criterion} relative to an ordered pair of variables \( (Q, A) \) in a DAG \( \mathcal{G} \) such that the following conditions hold: (1) \( Z_{\text{\rm CFD}} \) intercepts all directed paths from \( Q \) to \( A \); (2) there exists a set of variables $W$, called the conditioning variables of \( Z_{\text{\rm CFD}} \), such that all back-door paths from \( Q \) to \( Z_{\text{\rm CFD}} \) are blocked by \( W \); (3) all back-door paths from \( Z_{\text{\rm CFD}} \) to \( A \) are blocked by \(Q \cup W \).
\end{definition}

As illustrated in Figure~\ref{fig:c}, \( C \) satisfies all conditions of the conditional front-door criterion relative to \( (Q, A) \), where the external knowledge \( E \) serves as the conditioning variable of \( C \). Therefore, \( C \) can be used as a valid conditional front-door adjustment variable to identify the causal effect of \( Q \) on \( A \). 

We now apply Theorem~\ref{do} to derive \( P(A \mid \text{do}(Q)) \). The derivation proceeds as follows:
\begin{align}
	\centering
	&P(A | do(Q)) = \sum_{c}P(c| do(Q)) P(A| c,do(Q)) \nonumber\\
	&~= \sum_{c}P(c| do(Q))\sum_{e} P(A | \mathrm{do}(Q),c,e) P(e| \mathrm{do}(Q),c) \nonumber\\
	&~= \sum_{c}P(c| do(Q))\sum_{e} P(A | \mathrm{do}(Q),\mathrm{do}(c),e) P(e| \mathrm{do}(Q),c), \nonumber \\
	&\quad \text{since }  (A \indep C| Q, E) \text{ in } \mathcal{G}_{\overline{Q}\underline{C}}~\text{(Rule 2 in Theorem~\ref{do})} \nonumber\\
	&~= \sum_{c}P(c| do(Q))\sum_{e} P(A | \mathrm{do}(c),e) P(e| \mathrm{do}(Q),c), \nonumber\\
	&\quad \text{since }  (A \indep Q | C, E) \text{ in } \mathcal{G}_{\overline{C}\overline{Q(E)}}~\text{(Rule 3 in Theorem~\ref{do})} \nonumber\\
	&~= \sum_{c}P(c| do(Q))\sum_{e,~q} P(A | \mathrm{do}(c),q,e) P(q | \mathrm{do}(c),e) P(e| \mathrm{do}(Q),c), \nonumber\\
	&~= \sum_{c}P(c| do(Q))\sum_{e,~q} P(A | c,q,e) P(q | \mathrm{do}(c),e) P(e| \mathrm{do}(Q),c), \nonumber \\
	&\quad \text{since }  (A \indep C | Q, E) \text{ in } \mathcal{G}_{\underline{C}}~\text{(Rule 2 in Theorem~\ref{do})} \nonumber\\
	&~= \sum_{c}P(c| do(Q))\sum_{e,~q} P(A | c,q,e) P(q | e) P(e| \mathrm{do}(Q),c), \nonumber \\
	&\quad\text{since }  (Q \indep C | E) \text{ in } \mathcal{G}_{\overline{C(E)}}~\text{(Rule 3 in Theorem~\ref{do})}\nonumber\\
	&~= \sum_{c}P(c| do(Q))\sum_{e,~q} P(A | c,q,e) P(q | e) \frac{P(e, c | do(Q))}{P(c | \mathrm{do}(Q))}, \nonumber \\
	&\quad \text{since the chain rule of conditional probability}\nonumber\\
	&~= \sum_{c}P(c| do(Q))\sum_{e,~q} P(A | c,q,e) P(q | e) \frac{P(c | Q, e) p(e)}{P(c | \mathrm{do}(Q))} \nonumber \\
	&~= {\sum_{c,~e} P(c | Q, e)} {\sum_{q,~e} P(A | c,q,e) P(q | e) P(e)} \nonumber
\end{align}

In our setting, \( Q \) denotes the query, which remains fixed throughout the reasoning process. That is, we do not perform any intervention over \( Q \), and thus it can be treated as a constant rather than a random variable. Consequently, the term \( P(q \mid e) \) and the summation over \( q \) can be omitted, simplifying the expression for the causal effect as follows,
\begin{equation}
	\label{final}
	\centering
	P(A \mid do(Q)) = \sum_{c,~e}  \underbrace{P(c \mid Q, e)}_{\text{\ding{173}}} \underbrace{P(A \mid c,q,e)}_{\text{\ding{174}}}\underbrace{P(e)}_{\text{\ding{172}}}
\end{equation}

We begin by introducing the procedure for generating counterfactual external knowledge, which is a key step in estimating causal effects via conditional front-door adjustment. We then present Equation~\ref{final}, which decomposes the causal effect from \( Q \) to \( A \) into three components, each of which can be estimated independently to recover the overall causal effect. Components~\ding{172}, \ding{173}, and \ding{174} are detailed in Sections~\ref{1}, \ref{2}, and \ref{3}, respectively.

\subsection{Constructing Counterfactual External Knowledge}
\label{1}
In many LLM-based reasoning tasks, the query $Q$ is fixed and not subject to direct causal intervention. This poses a fundamental challenge for causal effect estimation. Traditional causal inference assumes the ability to intervene on the treatment, i.e., $Q$ in our case. However, in practice, we typically only observe outputs conditioned on a single realisation of $Q$. As a result of this limitation, directly estimating the causal effect $P(A \mid do(Q))$ becomes infeasible under standard assumptions.

To overcome this limitation, we introduce counterfactual external knowledge as contextual background for reasoning. While \( Q \) remains fixed, modifying \( E \) allows us to simulate alternative reasoning environments. These changes induce meaningful shifts in the distribution of \( C \), effectively mimicking how \( Q \) would behave under different contexts. In this way, we simulate causal intervention without directly altering \( Q \).

To simulate the causal intervention, we construct counterfactual versions of the external knowledge that alter the context in which reasoning occurs. Concretely, given \( E \), we first prompt the LLM to identify the top \( T \) entities most relevant to \( Q \), denoted as \( V = [v_1, v_2, \dots, v_T] \). These entities are ranked by their relevance scores and assigned corresponding weights \( W = [w_1, w_2, \dots, w_T] \), where \( w_1 \geq w_2 \geq \dots \geq w_T \). Then, for each \( v \in V \), we generate a counterfactual alternative \( v^* \), forming the set \( V^* = [v_1^*, v_2^*, \dots, v_T^*] \). The weights of the counterfactual entities are preserved, i.e., \( w_t \) remains unchanged, allowing a controlled substitution in the reasoning context. Here, the subscript \( t \in \{1, 2, \dots, T\} \) indexes the entities based on their ranked relevance to \( Q \). 

Next, from the counterfactual entity list \( V^* \), we enumerate all possible subsets of size \( T-1 \), resulting in \( \binom{T}{T-1} = T \) combinations. For each combination, we replace the corresponding entities in the original external knowledge \( e \) to construct a counterfactual variant \( e^* \), as follows:
\begin{equation}
	\centering
	E^* = [e_1^*, e_2^*, \dots, e_T^*],
\end{equation}
where \( t \in \{1, 2, \dots, T\} \) indexes the combinations, each formed by removing exactly one entity from the original set of \( T \) counterfactual entities.

We assign a probability to each \( e_t^* \) based on the product of the weights of its constituent entities. Let \( W_t = \{w_{t,1}, \dots, w_{t,T-1}\} \) be the weights of the entities in the \( t \)-th combination. Here, each \( W_t \) contains the weights of the \( T-1 \) entities, since each \( e_t^* \) is constructed by removing exactly one entity from the original set of \( T \) counterfactual entities. Then, the probability of \( e_t^* \) is defined as follows:
\begin{equation}
	\label{111}
	\centering
	P(e_t^*) = \frac{\prod_{i=1}^{T-1} w_{t,i}}{\sum_{t=1}^{T} (\prod_{i=1}^{T-1} w_{t,i})}.
\end{equation}

\subsection{The calculation of \( P(c \mid Q, e) \)}
\label{2}
We prompt the LLM to generate \( M \) CoTs based on the query \( Q \) and external knowledge \( E \). These CoTs are then encoded into vector representations using a dedicated encoder. We then apply the K-means clustering algorithm~\cite{DBLP:journals/isci/IkotunEAAJ23} to partition the \( M \) CoTs into \( N \) clusters. The CoT closest to each cluster centroid is selected, resulting in \( N \) representative CoTs for downstream causal interventions.

Since the encoder operates in a representation space different from that of the LLM~\cite{DBLP:journals/corr/abs-2403-02738}, the computed distances between CoTs may not accurately reflect the LLM's reasoning preferences. To mitigate this issue, we adopt contrastive learning~\cite{DBLP:conf/icml/ChenK0H20,DBLP:conf/emnlp/Liang00ZTC22} to fine-tune the encoder so that its embedding space is aligned with the LLM representation space.

Specifically, we construct a training dataset \(\mathcal{D}\), where each instance consists of \( Q \) and its associated CoT \( c \). We treat each CoT \( c \in \mathcal{D} \) as an anchor, and denote its representation as \( z_a \). Then, we prompt the LLM to generate a semantically similar CoT \( c^+ \), whose representation is used as the positive sample \( z^+ \). Meanwhile, CoTs \( c^- \) from other instances in the same batch serve as negative samples, with their representations denoted as \( z^- \). Following prior works~\cite{DBLP:conf/icml/ChenK0H20,DBLP:journals/corr/abs-2403-02738}, we adopt the InfoNCE loss to fine-tune the encoder:
\begin{equation}
	\centering
	\mathcal{L} = -\log \frac{\exp(z_a^\top z^+ / \tau)}{\exp(z_a^\top z^+ / \tau) + \sum_{z^-} \exp(z_a^\top z^- / \tau)},
\end{equation}where \( \tau \) is a temperature parameter that controls the scaling of the similarities, and \( \sum_{z^-} \) denotes the summation over all negative sample embeddings in the batch.

For each counterfactual external knowledge \( e_t^* \in E^* \), we input it together with \( Q \) into the LLM and repeat this process \( P \) times to generate CoTs. These generated CoTs form the set $C_t^* = [c^*_{t,1}, c^*_{t,1}, \dots, c^*_{t,p}]$. We then encode each CoT in this set using the encoder, resulting in the representations $\tilde{C}_t^* = [\tilde{c}^*_{t,1}, \tilde{c}^*_{t,2}, \dots, \tilde{c}^*_{t,p}]$. Similarly, the CoT \( C \) is encoded into the representations \( \tilde{C} \).

Next, we use cosine similarity to compute the similarity between \( \tilde{c}_n \), the representation of the \( n \)-th CoT selected from the \( N \) clusters, and each representation in \( \tilde{C}_t^* \), as follows:
\begin{equation}
	\centering
	d_{n,t,p} = \text{cosine}(\tilde{c}_n, \tilde{c}^*_{t,p}),
\end{equation} where \( d_j \in [-1, 1] \). A value close to 1 indicates high semantic similarity, while a value near 0 suggests dissimilarity.

To assess the impact of the counterfactual external knowledge on the stability of reasoning, we define a similarity threshold \( s \). For each generated CoT \( c^*_{t,p} \) from set \( C^*_t \), we calculate the similarity score \( d_{n,t,p} \). If \( d_{n,t,p} \geq s \), we consider \( c_{t,p} \) to be logically consistent with the original CoT \( c_n \), assigning it a score of 1. Conversely, if \( d_{n,t,p} < s \), we deem it to differ significantly from \( c_n \), and assign it a score of 0. Formally, we define the indicator function as:
\begin{equation}
	\centering
	\mathbb{I}_{\text{sim}}(c_n,c^*_{t,p}) = 
	\begin{cases} 
		1 & \text{if } d_{n,t,p} \geq s, \\
		0 & \text{if } d_{n,t,p} < s.
	\end{cases}
\end{equation}

We then compute the conditional probability \( P(c \mid Q, e) \) by averaging the indicator scores over the \( P \) generated CoTs:
\begin{equation}
	\label{222}
	\centering
	P(c \mid Q, e) \approx \frac{1}{P} \sum_{p=1}^P \mathbb{I}_{\text{sim}}(c_n,c^*_{t,p}).
\end{equation}

This probability reflects the proportion of generated CoTs that remain semantically consistent with the original CoT under the given \( E \) and \( Q \).

\subsection{The calculation of \( P(A \mid c,q,e) \)}
\label{3}
Through Section~\ref{1} and~\ref{2}, we identify CoTs generated using counterfactual external knowledge that are semantically consistent with the original CoTs, forming the subset:
\begin{equation}
	\centering
	C_{t,\text{sub}}^* = [c^*_{t,1}, c^*_{t,1}, \dots, c^*_{t,r}]
\end{equation}
where $C_{t,\text{sub}}^* \subseteq C_t^*$ and \( r \in \{1, 2, \dots, R\} \). $R$ is the number of CoTs that satisfy \( d_{n,t,p} > s \), indicating semantic consistency with \( c_n \).

For each \( c^*_{t,r} \in C_{t,\text{sub}}^* \), we compute the answer using a reasoning function:
\begin{equation}
	\centering
	a_{t,r} = f(\tilde{c}^*_{t,r}),
\end{equation}
where \( f(\cdot) \) denotes the LLM-based reasoning process that produces an answer given a CoT representation.

We compare \(  a_{t,r} \) against the reference answer \( a_n \) (i.e, $a_{n} = f(\tilde{c}_{n})$). If \(  a_{t,r} = a_n \), it suggests that \( {c}^*_{t,r} \) is insensitive to external knowledge changes, and we assign a score of 0. Otherwise, we assign a score of 1, indicating that \( {c}^*_{t,r} \) reflects adaptive reasoning. We formalise this using an indicator function:
\begin{equation}
	\centering
	\mathbb{I}_{\text{ins}}({c}^*_{t,r}) = 
	\begin{cases} 
		0 & \text{if } a_{t,r} = a_{n}, \\
		1 & \text{if } a_{t,r} \neq a_{n}.
	\end{cases}
\end{equation}

The probability \( P(A \mid c, q, e) \) is defined as:
\begin{equation}
	\label{333}
	\centering
	P(A \mid c, q, e) \approx \frac{1}{R} \sum_{j=1}^R \mathbb{I}_{\text{ins}}({c}^*_{t,r}),
\end{equation}

This equation captures the sensitivity of the CoT \( c_i \) to the external knowledge variations. A low value of \( P(A \mid c, q, e) \) indicates consistent answers across contexts, while a high value suggests that the reasoning outcome is highly responsive to external knowledge.

\subsection{The calculation of \(P(A \mid do(Q)) \)}
Following the previous derivation, Equation~\ref{final} can be simplified to the following form:
\begin{align}
	\label{666}
	\centering
	&P(A \mid do(Q)) = \sum_{c,~e}  {P(c \mid Q, e)} {P(A \mid c,q,e) P(e)} \notag \\ 
	&\quad \approx \sum_{n=1}^{N} \sum_{t=1}^{T} \left[ \frac{1}{P} \sum_{j=1}^{P} \mathbb{I}_{\text{sim}}(c_n,c^*_{t,p}) 
	\cdot \frac{1}{R} \sum_{j=1}^{R} \mathbb{I}_{\text{ins}}({c}^*_{t,r})  \cdot P(e_t^*) \right]
\end{align}

\section{Experiments}
In this section, we evaluate the effectiveness and robustness of CFD-Prompting across four knowledge-intensive datasets using three different backbone LLMs. We first introduce the datasets and baseline methods used for comparison, followed by implementation details. We then present the main results, conduct robustness and hyper-parameter studies to assess stability under noisy conditions, and perform an ablation study to examine the contribution of key components in our framework. Due to space constraints, the case study illustrating the practical application of our framework is provided via the anonymous link in the abstract.

\subsection{Dataset and Evaluation}
To better evaluate the performance of our framework in handling complex knowledge-intensive tasks~\cite{DBLP:journals/corr/abs-2009-02252,DBLP:journals/corr/abs-2309-12767}, we follow previous works~\cite{DBLP:conf/acl/Wu0CWRKRM24,DBLP:journals/corr/abs-2403-02738} and select the SciQ, HotpotQA, WikiHop, and MuSiQue datasets for evaluation. These four datasets cover diverse knowledge domains~\cite{DBLP:conf/aclnut/WelblLG17,DBLP:conf/emnlp/Yang0ZBCSM18,DBLP:journals/tacl/WelblSR18,DBLP:journals/tacl/TrivediBKS22}, include multi-hop reasoning and feature various question types. They comprehensively assess the performance of methods in knowledge retrieval, reasoning ability, and information processing. The detailed information for each dataset are as follows:
\begin{itemize}[leftmargin=0.4cm]
	\item \textbf{SciQ}~\cite{DBLP:conf/aclnut/WelblLG17} is a multiple-choice science QA dataset covering physics, chemistry, and biology. We evaluate comparison methods and ours on the test set with provided supporting evidence.
	\item \textbf{HotpotQA}~\cite{DBLP:conf/emnlp/Yang0ZBCSM18} is a multi-hop QA benchmark with open-ended and yes/no questions, requiring reasoning across multiple supporting documents. We use the provided documents as external knowledge in our experiments.
	\item \textbf{WikiHop}~\cite{DBLP:journals/tacl/WelblSR18} is a multi-choice, multi-hop reasoning dataset. We treat its queries as questions and prompt models to generate free-form answers instead of selecting from candidates.
	\item \textbf{MuSiQue}~\cite{DBLP:journals/tacl/TrivediBKS22} emphasises multi-step reasoning and compositional question decomposition. We select instances requiring more than three reasoning hops for evaluation.
\end{itemize}

We use Exact Match (EM) and F1 score as the evaluation metrics to assess method performance~\cite{DBLP:conf/emnlp/Yang0ZBCSM18}. Following previous work~\cite{DBLP:conf/ijcnlp/LyuHSZRWAC23}, we extract the text span immediately following the keyword ``answer is'' as the final predicted answer.

\begin{table*}[t]
	\setlength\tabcolsep{7pt}
	\centering
	\caption{The comparison results of CFD-Prompting and seven methods across three backbone LLMs on four knowledge-intensive tasks. Best results are highlighted in bold.}
	\label{tab:main}
	\begin{tabular}{c|c|cc|cc|cc|cc|cc}
		\toprule
		& & \multicolumn{2}{c}{\textbf{SciQ}} & \multicolumn{2}{c}{\textbf{HotpotQA}} & \multicolumn{2}{c}{\textbf{WikiHop}} & \multicolumn{2}{c|}{\textbf{MuSiQue}} & \multicolumn{2}{c}{\textbf{Average}}\\ \cmidrule{3-12}
		\textbf{Model}& \textbf{Method}& EM~$\uparrow$ & F1~$\uparrow$ & EM~$\uparrow$ & F1~$\uparrow$ & EM~$\uparrow$ & F1~$\uparrow$ & EM~$\uparrow$ & F1~$\uparrow$ & EM~$\uparrow$ & F1~$\uparrow$\\
		\midrule
		\multirow{7}{*}{LLaMA-2} 
		& ICL         & 7.81  & 9.56  & 8.40  & 11.95 & 11.60 & 15.47 & 1.33  & 2.85  & 7.29  & 9.96 \\
		& CoT w/o ctx & 14.72 & 22.63 & 8.20  & 14.23 & 4.40  & 6.26  & 0.40  & 3.07  & 6.93  & 11.55 \\
		& CoT         & 30.55 & 45.98 & 9.50  & 17.19 & 16.10 & 21.22 & 1.73  & 4.50  & 14.47 & 22.22 \\
		& CoT-SC      & 41.63 & 54.23 & 17.20 & 24.46 & 21.30 & 26.78 & 2.13  & 4.62  & 20.56 & 27.52 \\
		& CAD         & 31.79 & 40.11 & 18.00 & 29.45 & 16.40 & 20.03 & 1.46  & 6.62  & 16.91 & 24.05 \\
		& DeCoT       & 42.26 & 54.43 & 20.40 & 32.16 & 19.59 & 25.92 & 3.32  & 6.27  & 21.39 & 29.70 \\
		& CP          & 42.10 & 54.23 & 17.90 & 25.54 & 22.01 & 29.03 & 2.99  & 8.05  & 21.25 & 29.21 \\\cmidrule{2-12}
		& Ours        & \textbf{43.60} & \textbf{55.35} & \textbf{22.00} & \textbf{33.73} & \textbf{22.93} & \textbf{31.06} & \textbf{4.52} & \textbf{11.68} & \textbf{23.26} & \textbf{32.95} \\
		\midrule
		\multirow{7}{*}{LLaMA-3} 
		& ICL         & 24.10 & 41.45 & 3.20  & 20.15 & 6.80  & 25.96 & 3.99  & 6.65  & 9.52  & 23.55 \\
		& CoT w/o ctx & 35.29 & 48.48 & 15.30 & 25.23 & 12.60 & 17.91 & 2.39  & 7.70  & 16.39 & 24.83 \\
		& CoT         & 50.32 & 67.52 & 31.30 & 47.85 & 22.40 & 32.27 & 12.63 & 20.15 & 29.16 & 41.95 \\
		& CoT-SC      & 60.86 & 77.57 & 36.30 & 54.80 & 26.60 & 37.12 & 21.48 & 31.20 & 36.31 & 50.17 \\
		& CAD         & 52.60 & 65.20 & 30.30 & 40.58 & 24.00 & 32.22 & 12.63 & 23.12 & 29.88 & 40.28 \\
		& DeCoT       & 62.18 & 79.33 & 43.30 & 59.51 & 24.25 & 35.17 & 21.14 & 28.85 & 37.72 & 50.71 \\
		& CP          & 61.59 & 77.30 & 42.74 & 58.87 & 25.20 & 35.22 & 22.13 & 24.18 & 37.91 & 48.89 \\\cmidrule{2-12}
		& Ours        & \textbf{63.12} & \textbf{79.65} & \textbf{47.80} & \textbf{62.41} & \textbf{27.20} & \textbf{37.55} & \textbf{24.35} & \textbf{34.12} & \textbf{40.62} & \textbf{53.43} \\
		\midrule
		\multirow{7}{*}{GPT-3.5 Turbo} 
		& ICL         & 65.95 & 81.01 & 41.20 & 52.13 & 21.30 & 31.43 & 23.01 & 33.20 & 37.87 & 49.44 \\
		& CoT w/o ctx & 42.19 & 55.58 & 30.30 & 42.89 & 16.30 & 23.26 & 8.91  & 17.65 & 24.42 & 34.84 \\
		& CoT         & 66.97 & 80.32 & 43.90 & 60.30 & 26.72 & 36.20 & 26.20 & 36.40 & 40.95 & 53.30 \\
		& CoT-SC      & 68.55 & 82.37 & 51.00 & 66.23 & 28.18 & 38.22 & 33.38 & 44.27 & 45.28 & 57.77 \\
		& CAD         & 67.08 & 78.84 & 45.00 & 60.85 & 27.70 & 37.70 & 26.86 & 40.14 & 41.66 & 54.38 \\
		& DeCoT       & 70.98 & 84.08 & 51.30 & 67.79 & 31.09 & 40.14 & 34.59 & 47.36 & 46.99 & 59.84 \\
		& CP          & 70.93 & 83.76 & 51.10 & 66.53 & 29.56 & 39.45 & 33.78 & 47.82 & 46.34 & 59.39 \\\cmidrule{2-12}
		& Ours        & \textbf{71.83} & \textbf{85.12} & \textbf{53.40} & \textbf{68.67} & \textbf{32.00} & \textbf{41.17} & \textbf{36.17} & \textbf{48.01} & \textbf{48.35} & \textbf{60.74} \\
		\bottomrule
	\end{tabular}
\end{table*}

\subsection{Comparison Methods and Backbone Models}
We compare our framework with the following methods:

\begin{itemize}[leftmargin=0.8cm]
	\item \textbf{In-Context Learning (ICL)}~\cite{DBLP:conf/nips/BrownMRSKDNSSAA20}: Prompt LLMs with a few demonstration examples consisting of only questions and their corresponding answers, without any intermediate reasoning or explanatory context.
	\item \textbf{CoT without context (CoT w/o ctx)}~\cite{DBLP:conf/nips/Wei0SBIXCLZ22}: Apply CoT prompting without providing any external context, generating reasoning purely based on the query itself.
	\item \textbf{CoT}~\cite{DBLP:conf/nips/Wei0SBIXCLZ22}: Prompt LLMs with demonstration examples containing detailed reasoning chains, guiding the model step-by-step through the thought process required to reach an answer.
	\item \textbf{CoT self-consistency (CoT-SC)}~\cite{DBLP:conf/iclr/0002WSLCNCZ23}: An extension of CoT prompting where LLMs generate multiple reasoning chains for a given query, and majority voting is used to determine the final answer.
	\item \textbf{Context-aware Decoding (CAD)}~\cite{DBLP:conf/naacl/ShiHLTZY24}: Improve LLM generation by comparing output distributions with and without external context to enhance reasoning reliability.
	\item \textbf{De-biasing CoT (DeCoT)}~\cite{DBLP:conf/acl/Wu0CWRKRM24}: Mitigate internal knowledge bias by using external knowledge as an instrumental variable to estimate the average causal effect of the CoT on the answer, enabling the selection of logically correct reasoning paths.
	\item \textbf{Causal Prompting (CP)}~\cite{DBLP:journals/corr/abs-2403-02738}: Estimates the causal effect of the query on the answer using the standard front-door adjustment, but it is primarily tailored for general reasoning tasks and does not account for the complexities of knowledge-intensive settings.
\end{itemize}


In our experiments, we select three pre-trained LLMs as backbone models to ensure diversity and comparability: Llama-2-7b-chat-hf (LLaMA-2)~\cite{DBLP:journals/corr/abs-2307-09288}, Meta-Llama-3-8B-Instruct (LLaMA-3)~\cite{DBLP:journals/corr/abs-2402-16048}, and GPT-3.5 Turbo~\cite{BrownMRSKDNSSAA20}. These LLMs differ in terms of parameter scale, training strategies, and open-source versus closed-source design, providing a comprehensive foundation for evaluation.

\subsection{Implementation Details}
We deploy the LLM using the vLLM framework~\cite{DBLP:conf/sosp/KwonLZ0ZY0ZS23}. Compared with traditional Transformer serving frameworks, vLLM achieves higher throughput and faster response speed by leveraging optimised dynamic batching and efficient KV-cache management. In our framework, we first generate \( M = 30 \) initial CoTs, which are then clustered into \( N = 5 \) groups. To construct counterfactual external knowledge, we extract \( T = 5 \) entities from the original context for replacement during counterfactual generation.

\subsection{Main Results}
Table~\ref{tab:main} reports the performance of our proposed framework, CFD-Prompting, across three LLM backbones and seven comparison methods on four knowledge-intensive tasks. As the model size increases from LLaMA-2 to GPT-3.5 Turbo, LLMs demonstrate stronger reasoning capabilities, leading to overall improved performance across all methods. CFD-Prompting consistently outperforms all baselines under each backbone model. Notably, it achieves an average EM/F1 of 23.26/32.95 on LLaMA-2. On LLaMA-3, it yields 40.62/53.43, surpassing CP by $(+2.71)$ EM and $(+4.54)$ F1. On GPT-3.5 Turbo, CFD-Prompting reaches 48.35/60.74, setting new state-of-the-art results with consistent gains across datasets. These results demonstrate the strong generalisability and effectiveness of our framework across both smaller and larger LLMs.

Among the non-causality-based prompting baselines, CoT w/o ctx performs the worst due to its lack of external knowledge, which limits its capacity for multi-hop reasoning. ICL improves performance by providing in-context demonstrations, while CAD enhances answer generation by contrasting model outputs with and without external knowledge. CoT-SC extends standard CoT by aggregating multiple reasoning paths through majority voting, thereby mitigating sampling variance.

DeCoT addresses internal bias by using external knowledge as an instrumental variable to estimate the average causal effect of CoTs on answers. It treats a CoT as logically correct if its ACE is greater than zero. However, this binary threshold offers limited granularity. Our framework instead estimates the causal effect of \( Q \) on \( A \), and ranks candidate answers accordingly. While CP applies the standard front-door adjustment, it assumes no observed confounders influence both query and CoT, a condition often violated in knowledge-intensive tasks. In contrast, CFD-Prompting leverages the conditional front-door adjustment, which allows for observed confounders such as external knowledge. This yields more accurate de-biasing and leads to consistent performance improvements across all benchmarks.

\begin{table}[t]
	\setlength\tabcolsep{4pt}
	\centering
	\caption{The results of the robustness study on the SciQ using LLaMA-3. Best results are highlighted in bold.}
	\label{tab:robustness}
	\begin{tabular}{c|cc|cc|cc}
		\toprule
		& 
		\multicolumn{2}{c|}{\textbf{SciQ}} & 
		\multicolumn{2}{c|}{\textbf{SciQ-Injected}} & 
		\multicolumn{2}{c}{\textbf{SciQ-Shuffled}} \\ \cmidrule{2-7} 
		\textbf{Method}& {EM}~$\uparrow$ & {F1}~$\uparrow$ & {EM}~$\uparrow$ & {F1}~$\uparrow$ & {EM}~$\uparrow$ & {F1}~$\uparrow$ \\
		\midrule
		ICL & 24.10 & 41.45 & 18.33 & 34.97 & 19.46 & 35.76 \\
		CoT w/o ctx & 35.29 & 48.48 & 35.29 & 48.48  &  35.29 & 48.48  \\
		CoT      & 50.32 & 67.52 & 43.78 & 60.32 & 46.15 & 62.14 \\
		CoT-SC   & 60.86 & 77.57 & 56.74 & 74.32 & 58.59 & 75.70 \\
		CAD      & 52.60 & 65.20 & 51.47 & 63.28 & 48.98 & 60.83 \\
		DeCoT    & 62.18 & 79.33 & 61.60 & 78.84 & 61.53 & 78.00   \\
		CP   & 61.59    & 77.30    & 58.74    & 74.60    & 59.79    & 76.68    \\\midrule
		Ours     & \textbf{63.12} & \textbf{79.65} & \textbf{62.29} & \textbf{79.39} & \textbf{62.17} & \textbf{78.90} \\
		\bottomrule
	\end{tabular}
	\vspace{-0.25cm}
\end{table}

\subsection{Robustness Study}
To evaluate the effectiveness and stability of our framework under noisy conditions, we design a robustness experiment based on the SciQ dataset. We introduce two types of perturbations: (1) SciQ-Injected, where we inject irrelevant content (10\% of the total) into the support documents to simulate noise interference; and (2) SciQ-Shuffled, where we randomly shuffle half of the support sentences to assess the model’s adaptability to contextual disorder. We adopt LLaMA-3 as the backbone model and report the results in Table~\ref{tab:robustness}.

As shown in the table, all methods except CoT w/o ctx experience performance degradation under the perturbations. We observe that CoT w/o ctx maintains unchanged results, as it does not utilise external knowledge and is thus unaffected by modifications to the context. Both DeCoT and CFD-Prompting demonstrate notable robustness. This can be attributed to their use of counterfactual external knowledge construction, which effectively mitigates the impact of noisy or disordered information. Compared to DeCoT, which selects entities based on frequency heuristics~\cite{DBLP:conf/acl/Wu0CWRKRM24}, CFD-Prompting focuses on query-relevant entity selection combined with conditional front-door adjustment, resulting in more reliable performance. Specifically, CFD-Prompting achieves the highest F1 scores of 79.39 on SciQ-Injected and 78.90 on SciQ-Shuffled. Even under perturbations, the performance drop remains minimal, highlighting the robustness and stability of our framework.

\begin{table}[t]
	\setlength\tabcolsep{4pt}
	\centering
	\caption{The performance of CFD-Prompting with different numbers of CoTs (\(M\)) and clusters (\(N\)) across four knowledge-intensive tasks.}
	\label{tab:hyperparam}
	\begin{tabular}{c|cc|cc|cc|cc}
		\toprule
		& 
		\multicolumn{2}{c|}{\textbf{SciQ}} & 
		\multicolumn{2}{c|}{\textbf{HotpotQA}} & 
		\multicolumn{2}{c|}{\textbf{WikiHop}} & 
		\multicolumn{2}{c}{\textbf{MuSiQue}} \\ \cmidrule{2-9} 
		{M}& {EM}~$\uparrow$ & {F1}~$\uparrow$ & {EM}~$\uparrow$ & {F1}~$\uparrow$ & {EM}~$\uparrow$ & {F1}~$\uparrow$ & {EM}~$\uparrow$ & {F1}~$\uparrow$ \\
		\midrule
		10 & 60.75    & 74.77  & 45.75    & 61.20    & 25.63    & 35.29  & 22.77    & 31.98 \\
		20    & 62.55   & 76.36  & 46.83    & 62.38    & 27.10    & 36.85  & 23.50    & 32.76 \\
		30   & 63.12    & 79.65  & 47.80    & 62.41    & 27.20    & 37.55  & 24.35    & 34.12 \\
		40    & 62.13    &76.45  & 47.17    & 62.61    & 27.25    & 36.72  & 24.27    & 33.65 \\
		50   & 63.57    & 79.79  & 48.20    & 63.71    & 27.50    & 37.53  & 25.17    & 34.21 \\
		\bottomrule
		\toprule
		{N}& {EM}~$\uparrow$ & {F1}~$\uparrow$ & {EM}~$\uparrow$ & {F1}~$\uparrow$ & {EM}~$\uparrow$ & {F1}~$\uparrow$ & {EM}~$\uparrow$ & {F1}~$\uparrow$ \\
		\midrule
		1 & 60.63    & 74.64  & 46.19    & 60.46    &26.30    & 36.28  & 22.85    & 32.10 \\
		3    & 63.03    & 76.89  & 47.40    & 62.56    & 26.66    & 36.54  & 23.27    & 32.62 \\
		5    & 63.12    & 79.65  & 47.80    & 62.41    & 27.20    & 37.55  & 24.35    & 34.12 \\
		7    & 62.50   & 76.33  & 47.51    & 62.80    & 27.39    & 37.74  & 24.23  & 33.95 \\
		9   & 63.34    & 80.08  & 48.06    & 62.88    & 27.73 & 37.63  & 24.53    & 34.78 \\
		\bottomrule
	\end{tabular}
	\vspace{-0.25cm}
\end{table}

\subsection{Hyper-parameter Study}
We further study the influence of the number of initially generated CoTs (\( M \)) and the number of clustering categories (\( N \)) on the performance of our framework. Table~\ref{tab:hyperparam} summarises the experimental results, where the upper part reports the effect of varying \( M \) while fixing \( N=5 \), and the lower part shows the effect of varying \( N \) while fixing \( M=30 \). The results indicate that increasing \( M \) generally improves model performance, while a larger \( N \) enables finer-grained clustering, further enhancing the robustness of causal effect estimation. However, larger \( M \) and \( N \) values incur higher token consumption and inference costs. To balance efficiency and performance, we set \( M=30 \) and \( N=5 \) as the default configuration throughout all experiments.

\begin{table*}[t]
	\setlength\tabcolsep{8pt}
	\centering
	\caption{The results of ablation study on four knowledge-intensive tasks using LLaMA-3. Best results are highlighted in bold.}
	\label{tab:Ablation}
	\begin{tabular}{l|cc|cc|cc|cc|cc}
		\toprule
		& \multicolumn{2}{c}{\textbf{SciQ}} & \multicolumn{2}{c}{\textbf{HotpotQA}} & \multicolumn{2}{c}{\textbf{WikiHop}} & \multicolumn{2}{c|}{\textbf{MuSiQue}} & \multicolumn{2}{c}{\textbf{Average}} \\ \cmidrule{2-11}
		\textbf{Method}& EM~$\uparrow$ & F1~$\uparrow$ & EM~$\uparrow$ & F1~$\uparrow$ & EM~$\uparrow$ & F1~$\uparrow$ & EM~$\uparrow$ & F1~$\uparrow$ & EM~$\uparrow$ & F1~$\uparrow$\\
		\midrule
		CFD-Promoting       & \textbf{63.12} & \textbf{79.65} & \textbf{47.80} & \textbf{62.41} & \textbf{27.20} & \textbf{37.55} & \textbf{24.35} & \textbf{34.12} & \textbf{40.62} & \textbf{53.43}\\
		\quad w Random weighting   & 60.57    & 77.87  & 44.80    & 60.06    & 26.30   & 36.28  & 22.50    & 31.45 & 38.54 & 51.42\\
		\quad w Reversed weighting   & 60.18    &76.43   & 40.85    &56.11     & 25.63   & 35.82  &21.93     &30.80 & 37.15 & 49.79\\
		\quad w/o Contrastive learning    & 62.24 & 79.33 & 45.98    & 60.91    & 26.49    & 36.54  & 23.27    & 31.74 & 39.50 & 52.13\\
		\quad w/o K-means clustering        & 62.47 & 79.25 & 46.12    & 60.92    & 26.51    & 36.84  & 23.64    & 32.07 & 39.69 & 52.27\\
		\bottomrule
	\end{tabular}
\end{table*}

\subsection{Ablation Study}
We conduct an ablation study to assess the contribution of three key components in CFD-Prompting: (1) relevance-based entity weighting during counterfactual construction, (2) encoder fine-tuning via contrastive learning, and (3) CoT clustering using K-means. All experiments are run on the LLaMA-3 model across four knowledge-intensive datasets. For (1), we compare our default relevance-based selection with two variants: (a) random selection and (b) reversed-weight selection, where the least relevant entities are chosen. For (2), we ablate contrastive learning by removing encoder fine-tuning. For (3), we disable clustering by using the full set of CoTs without K-means. These settings allow us to isolate and quantify the impact of each design choice.

As shown in Table~\ref{tab:Ablation}, removing any of the three components in CFD-Prompting leads to consistent performance degradation across all four datasets, validating their importance. The most substantial drop occurs when the relevance-based weighting strategy is ablated. Specifically, replacing it with random selection reduces average F1 from 53.43 to 51.42 (–2.01), while reversed weighting further lowers it to 49.79 (–3.64). Contrastive learning also proves essential. Removing encoder fine-tuning reduces the average F1 to 52.13, with notable drops on MuSiQue (–2.38) and HotpotQA (–1.50), suggesting that enhancing the encoder’s ability to distinguish different CoT representations improves the accuracy of causal effect estimation. Finally, removing K-means clustering leads to moderate but consistent drops, indicating that promoting CoT diversity through clustering contributes to more stable and robust causal effect estimation.

\section{Related Work}
\subsection{LLMs for Knowledge-Intensive Tasks}
Knowledge-intensive tasks~\cite{DBLP:journals/corr/abs-2009-02252,DBLP:journals/tacl/WelblSR18} require large-scale models to effectively leverage external information during inference. A common solution is RAG~\cite{DBLP:conf/emnlp/HoshiMNTMTD23,DBLP:journals/corr/abs-2304-04358}, which retrieves relevant knowledge from external corpora based on the input and supplements the model’s internal reasoning process. Building on this idea, ICL~\cite{DBLP:journals/corr/abs-2212-14024} provides input-output exemplars directly within the prompt, allowing the model to generalise task-specific behaviours from observed patterns. To further improve reasoning quality, CoT prompting~\cite{DBLP:conf/nips/Wei0SBIXCLZ22} encourages models to decompose complex problems into intermediate steps, enabling more systematic utilisation of retrieved or contextual information. Beyond prompting strategies, structured external resources such as knowledge graphs~\cite{DBLP:journals/dai/IbrahimAIK24,DBLP:conf/acl/0006WS24} and optimisation techniques like Reinforcement Learning from Human Feedback (RLHF)~\cite{DBLP:conf/iclr/DaiPSJXL0024} have also been developed to guide model inference and enhance the integration of external knowledge.

However, despite these advancements, large-scale models still exhibit internal bias that can distort their use of external information, often leading to flawed reasoning and incorrect answers.

\subsection{Chain-of-Thought}
In recent years, CoT prompting has been widely adopted to enhance the reasoning capabilities of LLMs. This technique has shown significant performance improvements in domains such as mathematical problem solving and symbolic reasoning~\cite{DBLP:conf/nips/Wei0SBIXCLZ22}. However, traditional CoT methods are prone to reasoning bias, where errors introduced in the early steps of a reasoning chain can propagate through subsequent steps, ultimately leading to biased final answers~\cite{valmeekam2022large}. This issue is particularly pronounced in knowledge-intensive tasks~\cite{DBLP:conf/emnlp/PressZMSSL23}, where longer reasoning chains and greater reliance on factual information increase the risk of compounding errors.

To address this, CoT-SC~\cite{DBLP:conf/iclr/0002WSLCNCZ23} generates multiple distinct reasoning paths and aggregates their outcomes to reduce the influence of individual erroneous paths. While effective in mitigating random inference errors, CoT-SC still suffers from internal bias in LLMs, particularly in tasks involving complex knowledge structures~\cite{DBLP:conf/ijcnlp/LyuHSZRWAC23}.

\subsection{De-biasing LLMs via Causal Inference}
Causal inference aims to quantify the effect of a treatment on an outcome~\cite{Pearl_2009,pearl2016causal}. Supported by a rich theoretical foundation, various methods have been developed to estimate causal effects even in the presence of unobserved confounders~\cite{icdm/ChengXXLLLZF23,10791303,hengXL0LL24,ChengXLLLGL24,ChengXLLLL23,DBLP:conf/pkdd/ChengXLLLL23}. Building on these foundations, an increasing number of studies apply causal inference to understand and mitigate biases in LLMs.

For example, \citet{li2025prompting} propose a causality-guided prompting framework to control the influence of social information on model predictions. Other works explore causal modelling for de-biasing tasks: \citet{WangCZCLLYLH22} design a causal graph for relation extraction and analyse counterfactual by removing textual context; \citet{ZhouMYYZ23} introduce Causal-Debias, which mitigates stereotypical associations via causal disentanglement; and \citet{WangMWZC23} develop a structural causal model to address entity bias through tractable interventions across entities, text, and outputs.

While these studies lay important groundwork, their scope is largely limited to fairness-oriented tasks or fine-tuning strategies. In contrast, recent causality-based prompting methods, such as DeCoT~\cite{DBLP:conf/acl/Wu0CWRKRM24}, CP~\cite{DBLP:journals/corr/abs-2403-02738}, and CAPITAL~\cite{abs-2507-00389}, aim to improve reasoning quality. DeCoT treats external knowledge as an instrumental variable to estimate the average causal effect of CoTs on answers, but offers only a coarse assessment. CP uses standard front-door adjustment, which relies on the strong assumption that no observed confounders exist between the prompt and the CoT, a condition often violated in knowledge-intensive tasks.

In contrast, the proposed CFD-Prompting estimates the causal effect of the query on the answer via conditional front-door adjustment, allowing for finer-grained de-biasing and delivering consistent state-of-the-art performance across benchmarks.

\section{Conclusion}
In this paper, we propose CFD-Prompting, a novel framework that leverages conditional front-door adjustment to mitigate internal bias in LLMs. CFD-Prompting constructs counterfactual external knowledge and aligns reasoning representations via contrastive learning, enabling more accurate estimation of the causal effect between the query and the answer. Unlike existing methods, CFD-Prompting relaxes restrictive assumptions and does not require access to model logits, making it applicable to both open- and closed-source LLMs. Extensive experiments on multiple knowledge-intensive benchmarks demonstrate that CFD-Prompting consistently improves reasoning accuracy and robustness, highlighting its effectiveness and generalisability in real-world applications.

\begin{acks}
This research was supported by the National Key Research and Development Program of China (Grant 2023YFF1000100), the Hubei Key Research and Development Program of China (Grants 2024BBB-055, 2024BAA008), the Major Science and Technology Project of Yunnan Province (Grant 202502AE090003), the Fundamental Research Funds for the Chinese Central Universities (Grant 2662025XX-PY005), and the research support package from the School of Computing Technologies at RMIT University.
\end{acks}

\section*{GenAI Usage Disclosure}
We used ChatGPT (OpenAI) solely for language polishing purposes during the preparation of this manuscript. No part of the research design, data analysis, code implementation, or scientific content generation involved the use of generative AI tools. All technical ideas, experimental results, and interpretations were produced independently by the authors.

\bibliographystyle{ACM-Reference-Format}
\balance
\bibliography{CFD-Prompting.bib}

\end{document}